\documentclass{article}

\usepackage{microtype}
\usepackage{graphicx}
\usepackage{subcaption}
\usepackage{booktabs} 

\usepackage{hyperref}

\usepackage[preprint]{icml2026}

\usepackage{amsmath}
\usepackage{amssymb}
\usepackage{mathtools}
\usepackage{amsthm}

\usepackage[capitalize,noabbrev]{cleveref}

\theoremstyle{plain}

\theoremstyle{definition}

\theoremstyle{remark}

\usepackage[textsize=tiny]{todonotes}

\icmltitlerunning{Causal Intervention-Based Memory Selection for Long-Horizon LLM Agents}

\begin{document}

\twocolumn[
  \icmltitle{Causal Intervention-Based Memory Selection \\ for Long-Horizon LLM Agents}

  \begin{icmlauthorlist}
    \icmlauthor{Saksham Sahai Srivastava}{uga}
  \end{icmlauthorlist}

  \icmlaffiliation{uga}{School of Computing, University of Georgia, Athens, Georgia, USA}

  \icmlcorrespondingauthor{Saksham Sahai Srivastava}{saksham.srivastava@uga.edu}

  \icmlkeywords{Large Language Models, Memory, Causal Inference, LLM Agents}

  \vskip 0.3in
]



\printAffiliationsAndNotice{}  

\begin{abstract}
Long-horizon LLM agents rely on persistent memory to support interactions across sessions, yet existing memory systems often retrieve context using semantic similarity or broad history inclusion, treating retrieved memories as uniformly useful. This assumption is fragile because memories may be topically related while remaining irrelevant, stale, or misleading. We propose \textit{Causal Memory Intervention} (CMI), a causal memory-selection technique that estimates how candidate memories affect the model's answer under controlled interventions, selecting memories that improve task performance while suppressing unstable, irrelevant, or harmful ones. To evaluate this setting, we introduce \textsc{Causal-LoCoMo}, a causally annotated benchmark derived from long conversational data, where each example contains a user request, a structured memory bank, useful memories, irrelevant distractors, and synthetic harmful memories. We compare CMI against vector, graph, reflection, summary, full-history, and no-memory baselines. Results show that CMI achieves a stronger balance between answer quality and robustness to misleading memory, suggesting that reliable long-term memory requires selecting context based on causal usefulness rather than relevance alone. The full framework, benchmark construction code, and experimental pipeline are available at \url{https://github.com/Saksham4796/causal-memory-intervention}.
\end{abstract}

\section{Introduction}

Large language model (LLM) agents are increasingly expected to operate beyond isolated, single-turn interactions. In realistic settings, an agent must remember user preferences, past decisions, unresolved goals, prior conversations, and domain-specific context accumulated over days, weeks, or months. This has motivated growing interest in long-term memory systems for LLM agents, where external memory stores are used to retrieve relevant information and condition future responses, as seen in Generative Agents~\cite{park2023generative}, MemoryBank~\cite{zhong2024memorybank}, MemGPT~\cite{packer2023memgpt}, Reflexion~\cite{shinn2023reflexion}, and Voyager~\cite{wang2023voyager}. Such memory mechanisms are central to personalized assistants, autonomous research agents, interactive tutoring systems, and long-horizon task-completion workflows, and are closely related to broader work on tool-using and interactive agents~\cite{yao2023react,schick2023toolformer,liu2024agentbench}. However, reliable memory use is not simply a matter of giving the agent more context. Existing memory systems often retrieve memories using semantic similarity~\cite{lewis2020retrieval,karpukhin2020dense}, summaries, graph neighborhoods, or broad conversation histories, implicitly assuming that retrieved memories are useful. In practice, this assumption is fragile. A memory can be topically related to the current request while being irrelevant, stale, ambiguous, semantically similar but contextually wrong, or even adversarially inserted. Prior work on long-context behavior~\cite{liu2024lost}, prompt injection~\cite{liu2023promptinjection}, RAG poisoning~\cite{zou2025poisonedrag}, and poisoned experience retrieval in MemoryGraft~\cite{srivastava2025memorygraft} suggests that models may overuse misleading context when it is placed in the input. When such memories are exposed to an LLM, the model may incorporate them into its response with high confidence, even when they should have been ignored. This creates a central challenge for long-horizon LLM agents: memory selection must distinguish memories that improve the current answer from memories that are merely relevant-looking or actively harmful.

Existing long-context and conversational-memory benchmarks typically evaluate whether a model can recover information from a long dialogue history. LoCoMo~\cite{maharana2024locomo} evaluates very long-term conversational memory, while LongBench~\cite{bai2024longbench}, L-Eval~\cite{an2024leval}, InfiniteBench~\cite{zhang2024infinitebench}, and RULER~\cite{hsieh2024ruler} evaluate long-context understanding across increasingly large input contexts. While valuable, this setup does not directly evaluate memory selection. If an agent answers correctly, it is often unclear whether the answer came from the right memory, from a spurious memory, from parametric knowledge, or from superficial lexical overlap. Similarly, when an agent fails, standard evaluation does not reveal whether the failure was caused by missing memory, poor retrieval, irrelevant distractors, or harmful memory contamination. This limitation is especially important for agentic systems, where the memory bank is not a passive transcript but an actively retrieved and curated source of evidence.

In this paper, we propose \textit{Causal Memory Intervention} (CMI), a causal memory-selection technique for long-horizon LLM agents. Rather than selecting memories only by semantic similarity~\cite{lewis2020retrieval,karpukhin2020dense} or compressed summaries, CMI estimates whether a candidate memory causally improves the agent's answer. This framing is inspired by causal intervention and counterfactual reasoning~\cite{pearl2009causality}, but applies intervention at the level of external memory rather than model internals. For each candidate memory, CMI compares model behavior under controlled intervention conditions, including no-memory, with-memory, and perturbed-memory settings. A memory is selected only when it improves the task score relative to the no-memory condition and remains stable under perturbation. This turns memory selection into an intervention-based decision problem: the goal is not to retrieve all relevant-looking memories, but to select memories whose presence has a positive causal effect on the current response while suppressing memories that are neutral, unstable, or harmful. This is complementary to prior causal analysis of model behavior~\cite{vig2020investigating,meng2022locating,geiger2025causal}, but focuses on the external memory context as the intervention surface.

To evaluate this idea, we construct \textsc{Causal-LoCoMo}, a causally annotated memory-selection benchmark derived from LoCoMo~\cite{maharana2024locomo}. Each example contains a current user request, a structured memory bank, useful memories, irrelevant distractors, and, when applicable, synthetic harmful memories designed to test robustness against misleading retrieval. Unlike standard long-conversation QA, \textsc{Causal-LoCoMo} is designed to test whether an agent can choose the memories that should influence the answer while avoiding memories that are merely semantically plausible or causally misleading. The benchmark includes temporal reasoning, multi-evidence questions, factual memory questions, and inferential memory questions. We further apply deterministic quality control and schema validation to ensure that the final examples support controlled comparison across memory-selection methods. We compare CMI against several representative memory strategies for LLM agents, including vector memory~\cite{lewis2020retrieval,karpukhin2020dense}, graph memory, reflection memory~\cite{shinn2023reflexion}, summary memory, full-history prompting~\cite{maharana2024locomo}, and no-memory prompting. Our results show that CMI achieves the strongest overall balance between answer quality and safe memory selection. Across 87 filtered \textsc{Causal-LoCoMo} examples, CMI obtains the best task score ($0.846$), highest success rate ($0.816$), and strongest useful-memory F1 ($0.875$). At the same time, it achieves near-perfect harmful-memory rejection ($0.990$) and zero poisoned-memory adoption. These results suggest that effective long-term memory requires more than retrieving or compressing context: agents must select memories according to their causal effect on the current answer.

This work makes the following contributions:
\begin{itemize}
    \item We introduce \textit{Causal Memory Intervention} (CMI), a causal memory-selection technique that selects memories based on their estimated intervention effect on the current answer.
    \item We construct \textsc{Causal-LoCoMo}, a causally annotated conversational-memory benchmark with useful, irrelevant, and harmful memory entries for evaluating long-horizon memory selection.
    \item We compare CMI against vector, graph, reflection, summary, full-history, and no-memory baselines using both answer-quality and memory-selection metrics.
    \item We empirically show that CMI improves the tradeoff between task performance and robustness, achieving strong useful-memory selection while substantially reducing harmful-memory adoption.
\end{itemize}

Overall, our results indicate that reliable LLM-agent memory requires causal discrimination: the ability to determine which memories should influence the current response and which should be ignored. CMI provides a practical step toward this goal by reframing memory selection as an intervention-based problem~\cite{pearl2009causality} rather than a purely similarity-based retrieval problem~\cite{lewis2020retrieval,karpukhin2020dense}.

\section{Related Work}
\label{sec:related_work}

Recent work has increasingly moved from single-turn language modeling toward agents that act, remember, and adapt over extended interactions. Agentic systems such as ReAct~\cite{yao2023react} combine reasoning and action in interactive environments, while Toolformer~\cite{schick2023toolformer} and broader agent benchmarks such as AgentBench~\cite{liu2024agentbench} study how language models use tools and perform multi-step tasks. Within this setting, long-term memory has become a central mechanism for continuity and personalization. Generative Agents~\cite{park2023generative} store natural-language records of experiences and use reflection to synthesize higher-level behavioral traces; Reflexion~\cite{shinn2023reflexion} stores verbal feedback to improve subsequent trials; MemoryBank~\cite{zhong2024memorybank} introduces mechanisms for updating and recalling long-term memories; and MemGPT~\cite{packer2023memgpt} frames memory management as an operating-system-like process over limited context. Other agent systems, such as Voyager~\cite{wang2023voyager}, maintain reusable skill libraries to support lifelong exploration. These works demonstrate that memory can improve agent persistence and adaptation. However, they generally treat memory as a useful resource once retrieved, whereas our work studies the memory-selection problem directly: given a noisy memory bank, which memories should actually be used for the current task?

A parallel line of work evaluates whether models can use long inputs or recover information from extended interaction histories. LoCoMo~\cite{maharana2024locomo} studies very long-term conversational memory across multi-session dialogues and evaluates models on question answering, event summarization, and multimodal dialogue generation. LongBench~\cite{bai2024longbench}, L-Eval~\cite{an2024leval}, InfiniteBench~\cite{zhang2024infinitebench}, and RULER~\cite{hsieh2024ruler} evaluate long-document understanding, multi-document reasoning, synthetic retrieval, and long-context robustness. Prior analysis further shows that models may fail to use relevant information reliably even when it is placed in the context window, especially when the information appears in the middle of a long context~\cite{liu2024lost}. These benchmarks are important for measuring long-context access, but they usually evaluate whether the model can answer from a supplied context. In contrast, our work evaluates whether an agent can choose the right memories before answering. \textsc{Causal-LoCoMo} therefore converts long-conversation QA into a memory-selection setting in which useful, irrelevant, and harmful memories are explicitly separated.

External memory for LLM agents is closely related to retrieval-augmented generation (RAG), where generation is conditioned on retrieved non-parametric evidence~\cite{lewis2020retrieval}. Dense retrieval~\cite{karpukhin2020dense}, reader-generator architectures~\cite{izacard2021leveraging}, retrieval-augmented pretraining~\cite{guu2020realm}, and retrieval from large token databases~\cite{borgeaud2022retro} have improved knowledge-intensive language tasks by retrieving external context before generation. However, retrieval introduces a selection problem: retrieved evidence may be semantically similar to the query while still being incomplete, irrelevant, stale, or misleading. This issue becomes more severe in persistent agent memory, where memory banks accumulate over time and may contain multiple related but contextually incompatible records. Our work differs from standard retrieval-based memory selection by ranking memories according to estimated causal effect on the current answer, rather than semantic similarity alone.

The reliability of memory-augmented agents is also connected to robustness and security in retrieval-augmented systems. Prompt injection attacks show that LLM-integrated applications can be steered by malicious instructions embedded in external content~\cite{liu2023promptinjection}. RAG poisoning work shows that corrupted retrieved documents can alter model outputs and induce attacker-chosen behavior~\cite{zou2025poisonedrag}. Recent work on MemoryGraft further demonstrates that poisoned experience retrieval can persistently compromise LLM agents by implanting malicious but successful-looking experiences into long-term memory, causing later agents to imitate unsafe patterns when semantically similar tasks retrieve those memories~\cite{srivastava2025memorygraft}. These works motivate the need for memory mechanisms that do not merely retrieve relevant-looking context, but actively suppress memories that are likely to mislead the agent. \textsc{CMI} addresses this issue by testing whether a candidate memory improves the answer under intervention and by rejecting memories whose effects are unstable or harmful.

Our work is also related to causal approaches for analyzing model behavior. Causal inference provides a formal language for interventions and counterfactual reasoning~\cite{pearl2009causality}. In NLP and interpretability, causal mediation analysis has been used to identify components that contribute to model predictions~\cite{vig2020investigating}, while causal tracing and model editing methods intervene on activations or parameters to locate and modify factual associations~\cite{meng2022locating}. More recent work on causal abstraction provides a theoretical foundation for mechanistic interpretability~\cite{geiger2025causal}. \textsc{CMI} is complementary to these internal intervention methods. Instead of intervening on neurons, activations, or weights, we intervene on the external memory context supplied to an LLM agent. This is a practical intervention surface because memory banks are editable, inspectable, and selected at inference time.

Prior work shows that memory can improve agent continuity, retrieval can augment model knowledge, and long-context benchmarks can test whether models recover information from extended inputs. However, these lines of work do not directly answer the question studied here: \emph{which memories should an LLM agent select from a noisy persistent memory bank?} We propose \textsc{Causal Memory Intervention} as a memory-selection technique that estimates whether candidate memories causally improve the current answer, and we introduce \textsc{Causal-LoCoMo} to evaluate this selection problem under useful, irrelevant, and harmful memory conditions. This framing shifts the focus from memory availability to causal memory usefulness.

\section{Proposed Framework}
\label{sec:proposed_framework}

We propose \textit{Causal Memory Intervention} (\textsc{CMI}), a memory-selection technique for long-horizon LLM agents. The goal of \textsc{CMI} is to select from a persistent memory bank only those memories that are expected to improve the current response, while rejecting memories that are irrelevant, unstable, or harmful. Unlike standard retrieval-augmented memory methods, which typically rank memories by lexical or embedding-based similarity to the query~\cite{lewis2020retrieval,karpukhin2020dense}, \textsc{CMI} ranks memories by their estimated causal effect on the model's answer.

Let $x$ denote the current user request and let $\mathcal{M} = \{m_1,m_2,\ldots,m_n\}$ denote the external memory bank available to the agent. Each memory $m_i$ is a natural-language record derived from previous interactions, observations, summaries, or adversarially inserted content. A memory-selection method outputs a subset $\widehat{\mathcal{S}} \subseteq \mathcal{M}$, which is then provided to a response model $p_\theta$ to generate the final answer $y \sim p_\theta(y \mid x, \widehat{\mathcal{S}})$. The central problem is to choose $\widehat{\mathcal{S}}$ such that the selected memories improve the response to $x$ without introducing misleading context.

\begin{figure*}[t]
    \centering
    \includegraphics[width=\textwidth]{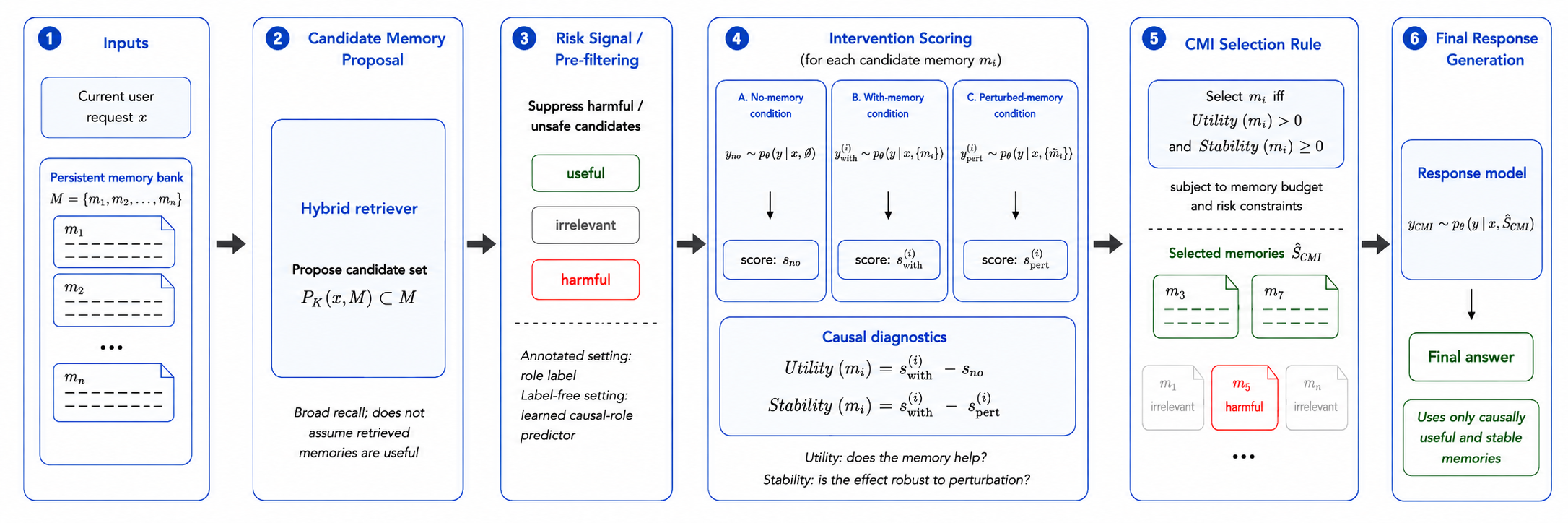}
    \caption{
    Overview of \textsc{Causal Memory Intervention}.
    }
    \label{fig:framework_diagram}
\end{figure*}

Figure~\ref{fig:framework_diagram} provides an overview of the proposed \textsc{CMI} pipeline. Given a current user request and a persistent memory bank, \textsc{CMI} first proposes candidate memories using a broad hybrid retriever. It then applies risk-aware filtering and intervention scoring by comparing no-memory, with-memory, and perturbed-memory conditions for each candidate. The final response is generated only from memories that satisfy the \textsc{CMI} selection rule.

\subsection{Memory Selection as a Causal Decision Problem}

A memory is useful not merely because it is semantically similar to the query, but because its inclusion changes the answer in a beneficial direction. We therefore model memory selection as an intervention problem. Given a candidate memory subset $\mathcal{S}$, we define the intervention $\operatorname{do}(\mathcal{C}=\mathcal{S})$ as the operation of forcing the response model to condition on $\mathcal{S}$ as its memory context. Let $Y_{\mathcal{S}} \sim p_\theta(y \mid x, \operatorname{do}(\mathcal{C}=\mathcal{S}))$ be the model response under this intervention. If $u(y,a^\star)\in[0,1]$ denotes a task utility measuring how well response $y$ satisfies the target answer or behavior $a^\star$, then the ideal value of a memory subset is
\[
    U(\mathcal{S})
    =
    \mathbb{E}_{Y_{\mathcal{S}}}
    \left[
        u(Y_{\mathcal{S}},a^\star)
    \right].
\]
The ideal memory selector would choose
\[
    \mathcal{S}^{\star}
    =
    \arg\max_{\mathcal{S}\subseteq \mathcal{M},\,|\mathcal{S}|\leq k}
    U(\mathcal{S}),
\]
where $k$ is the memory budget. Directly solving this problem is infeasible because evaluating all memory subsets is combinatorial. \textsc{CMI} therefore estimates the causal contribution of individual candidate memories through a small number of targeted interventions.

\subsection{Candidate Memory Proposal}

Given a task $x$ and memory bank $\mathcal{M}$, \textsc{CMI} first constructs a small candidate set $\mathcal{P}_K(x,\mathcal{M}) \subseteq \mathcal{M}$ using a hybrid retriever. This stage is intentionally broad: it proposes memories that may be relevant to the current task, but does not assume that all retrieved candidates are useful. The intervention stage then determines which of these candidates should actually be selected for final answer generation.

In the annotated benchmark setting used in our experiments, memories may also carry role information such as useful, irrelevant, or harmful. The implemented \textsc{CMI} agent uses this information as a risk signal to suppress memories marked harmful or otherwise unsafe before final selection. In a label-free deployment, this risk signal would be replaced by a learned or prompted causal-role predictor.

\subsection{Intervention Scoring}

For each candidate memory $m_i \in \mathcal{P}_K(x,\mathcal{M})$, \textsc{CMI} compares model behavior under three controlled conditions.

First, the model answers the task without memory: 
\[
    y_{\mathrm{no}} \sim p_\theta(y \mid x,\varnothing).
\]
Second, the model answers the task with candidate memory $m_i$:
\[
    y_{\mathrm{with}}^{(i)}
    \sim
    p_\theta(y \mid x,\{m_i\}).
\]
Third, the model answers the task with a perturbed version of the same memory:
\[
    y_{\mathrm{pert}}^{(i)}
    \sim
    p_\theta(y \mid x,\{\widetilde{m}_i\}),
\]
where $\widetilde{m}_i$ is obtained by perturbing the candidate memory. The perturbation is designed to test whether the apparent benefit of the memory is stable or whether the model is relying on brittle, misleading, or superficial cues.

Let $s(\cdot)$ denote the deterministic task scorer used inside the intervention loop. We compute
\[
\begin{aligned}
    s_{\mathrm{no}} &= s(y_{\mathrm{no}}, a^\star), \\
    s_{\mathrm{with}}^{(i)} &= s(y_{\mathrm{with}}^{(i)}, a^\star), \\
    s_{\mathrm{pert}}^{(i)} &= s(y_{\mathrm{pert}}^{(i)}, a^\star).
\end{aligned}
\]
The causal utility of memory $m_i$ is defined as
\[
    \operatorname{Utility}(m_i)
    =
    s_{\mathrm{with}}^{(i)}
    -
    s_{\mathrm{no}},
\]
and its stability is defined as
\[
    \operatorname{Stability}(m_i)
    =
    s_{\mathrm{with}}^{(i)}
    -
    s_{\mathrm{pert}}^{(i)}.
\]
Utility measures whether the memory improves the answer relative to having no memory. Stability measures whether this improvement is robust to perturbation. A memory with positive utility but negative stability may help only because of brittle or misleading phrasing, and is therefore not considered reliable.

\subsection{CMI Selection Rule}

The final \textsc{CMI} selection rule is:
\[
\begin{aligned}
m_i \in \widehat{\mathcal{S}}_{\textsc{CMI}}
\Longleftrightarrow
\begin{cases}
\operatorname{Utility}(m_i) > 0,\\
\operatorname{Stability}(m_i) \geq 0
\end{cases}
\end{aligned}
\]
subject to the memory budget and any risk-filtering constraints. Therefore, a memory is selected only if it improves the task score over the no-memory baseline and does not lose its effect under perturbation. The final response is then generated as:
\[
    y_{\textsc{CMI}}
    \sim
    p_\theta(y \mid x,\widehat{\mathcal{S}}_{\textsc{CMI}})
\]

This differs from conventional retrieval in two ways. First, selection is based on estimated change in task performance, not only on similarity between $x$ and $m_i$. Second, perturbation introduces a robustness check: a memory must not only help, but help in a stable way. As a result, \textsc{CMI} favors memories with positive causal influence and suppresses memories that are semantically plausible but non-causal or harmful.

\subsection{Comparison to Standard Memory Selection}

Standard retrieval-based memory selection can be written as:
\[
    \widehat{\mathcal{S}}_{\mathrm{retr}}
    =
    \operatorname{TopK}_{m_i\in\mathcal{M}}
    \operatorname{sim}(x,m_i)
\]
where $\operatorname{sim}(x,m_i)$ is typically an embedding or lexical similarity score. Such methods can retrieve memories that resemble the current query, but they do not directly test whether the memory improves the model's answer. In long-horizon memory banks, this distinction is important: harmful or irrelevant memories may share entities, topics, or temporal expressions with the query while still leading to the wrong answer.

\textsc{CMI} replaces similarity-based relevance with intervention-based usefulness. A memory is useful if the model performs better when the memory is present than when it is absent, and if that improvement is stable under perturbation. Thus, \textsc{CMI} treats memory selection as a causal decision problem~\cite{pearl2009causality} rather than a nearest-neighbor retrieval problem~\cite{karpukhin2020dense}.

This causal framing also determines how we compare memory-selection methods. A strong selector should not only improve final answer quality, but should do so by selecting memories with positive causal effect while rejecting irrelevant or harmful memories. We therefore evaluate \textsc{CMI} and all baselines using both task-level performance and memory-selection behavior, as detailed in Section~\ref{sec:experimental_setup}.

\section{Dataset}
\label{sec:dataset}

We construct \textsc{Causal-LoCoMo} from LoCoMo~\cite{maharana2024locomo} by converting evidence-grounded long-conversation question answering instances into causal memory-selection tasks. Each example is anchored in an original LoCoMo conversation, question, gold answer, and evidence annotation. The LoCoMo question is used as the current user request, while the original multi-session chat history is retained to support full-history baselines and to preserve the conversational setting from which memories are derived. The resulting benchmark is designed to compare memory-selection methods under a shared setting: each method receives the same current task and candidate memory bank, but differs in which memories it selects or exposes to the response model. For each task, we build a structured memory bank with three types of memories. First, evidence that supports the gold answer is rewritten into clear, self-contained useful memories. Second, realistic but non-answer-supporting memories are sampled from the same LoCoMo conversation and used as irrelevant distractors. Third, in selected examples, we add synthetic harmful memories that are plausible but misleading. These harmful memories typically contradict the gold answer, swap entities, introduce incorrect temporal information, or encourage an unsupported inference. For temporal questions, relative expressions such as \textit{yesterday}, \textit{last week}, or \textit{last year} are resolved using the corresponding session timestamp so that useful memories contain explicit and auditable temporal facts. Thus, each example directly tests whether a memory-selection method can identify memories that should influence the current answer while ignoring memories that are merely topically related or actively misleading.

The final benchmark is produced through an LLM-assisted construction pipeline followed by deterministic filtering and schema validation. We first generate 100 candidate \textsc{Causal-LoCoMo} examples from LoCoMo using GPT-5 as the dataset construction model. The deterministic filter removes examples whose useful memories do not sufficiently support the expected answer, whose scoring criteria are malformed or unreliable, or whose metadata violates the causal-memory schema. This yields 87 filtered evaluation examples. For experimental comparability, we retain the original LoCoMo chat histories after filtering, resulting in a final dataset with 432 past sessions and 491 memory entries: 89 useful memories, 348 irrelevant memories, and 54 harmful memories. The task distribution contains 46 temporal memory QA examples, 25 multi-evidence memory QA examples, 14 inferential memory QA examples, and 2 factual memory QA examples. This construction preserves the realism of long conversational memory while exposing the central question studied in this work: whether a memory-selection technique can choose memories that improve the current answer and suppress memories that are semantically plausible but non-causal or harmful.

\section{Experimental Setup}
\label{sec:experimental_setup}

We evaluate \textsc{CMI} as a causal intervention-based memory-selection technique for long-horizon LLM agents. Our goal is to compare \textsc{CMI} against alternative memory access strategies under the same task, memory bank, response model, and scoring pipeline. Each \textsc{Causal-LoCoMo} example contains a current user request, a structured memory bank, optional past conversational sessions, and scoring criteria. The memory bank contains useful, irrelevant, and harmful memories, while the past sessions preserve the original LoCoMo conversation history for full-history baselines. All methods share the same final response model and final scoring pipeline. The methods differ only in how they construct the context provided to the response model. In the case of \textsc{CMI}, this includes an additional pre-answer intervention procedure for selecting memories.

\subsection{Compared Methods}

We compare \textsc{CMI} against several memory access strategies for LLM agents.

\paragraph{No memory.}
The no-memory baseline receives only the current user request and no external memory. This measures how well the model can answer from the task wording and its parametric prior alone.

\paragraph{Full history.}
The full-history baseline receives the original LoCoMo~\cite{maharana2024locomo} past sessions associated with the current example. This represents a long-context agent that does not perform explicit memory selection. It tests whether directly providing the complete conversational history is sufficient for reliable answer generation. For memory-selection bookkeeping metrics, this baseline is treated as exposing the full memory bank, even though the actual prompt uses past sessions.

\paragraph{Summary memory.}
The summary-memory baseline summarizes the structured memory bank, rather than the raw past sessions. The resulting summary is inserted as a single synthetic summary memory and provided to the response model. For memory-selection bookkeeping metrics, this baseline is also treated as exposing the full memory bank, since the summary is derived from all available memories.

\paragraph{Vector memory.}
The vector-memory baseline embeds memory entries and retrieves the top memories by semantic similarity to the current user request. This represents a standard retrieval-augmented memory system in which similarity is used as the primary relevance signal.

\paragraph{Graph memory.}
The graph-memory baseline retrieves memories using a lightweight lexical and scope-based graph over the memory bank. In this graph, memory nodes are connected to scope nodes and token or concept nodes, and retrieval favors memories that are close to the current query in this graph. This baseline tests whether simple graph structure over memory text and memory scope improves selection over vector similarity alone.

\paragraph{Reflection memory.}
Inspired by reflection-style agent memory~\cite{shinn2023reflexion}, the reflection-memory baseline constructs deterministic reflections from memory entries. In the implemented version, these reflections are generated using each memory's label, for example converting useful memories into lesson-like reflections and harmful memories into safety-oriented reflections. The method then retrieves reflections using keyword-based matching. Thus, this baseline is not an LLM self-reflection method; rather, it is a label-aware reflection-style memory representation.

\paragraph{Causal Memory Intervention.}
\textsc{CMI} is our proposed memory-selection method. It first proposes candidate memories and then evaluates them through no-memory, with-memory, and perturbed-memory intervention conditions. Memories are selected when they improve the deterministic task score relative to the no-memory condition and remain stable under perturbation.

\subsection{Evaluation Pipeline}

For memory-bank methods, the selected memory context is passed through the same final memory prompt. The no-memory and full-history baselines use corresponding no-context and past-session prompts, respectively, while sharing the same response model and scoring pipeline. The model is instructed to answer the current task using the provided memory or history. The generated answer is then scored using the dataset-specific expected answer and scoring criteria. We use the same final scoring procedure for all methods so that reported performance is comparable across memory strategies.

The primary scalar score for each prediction is a hybrid task score:
\[
    s(y, a^\star)
    =
    0.7\,s_{\mathrm{det}}(y, a^\star)
    +
    0.3\,s_{\mathrm{judge}}(y, a^\star)
\]
where $s_{\mathrm{det}}$ is a deterministic scorer and $s_{\mathrm{judge}}$ is a GPT-5-based judge score. The deterministic scorer checks fields such as \texttt{must\_include}, \texttt{must\_not\_include}, \texttt{expected\_answer}, \texttt{max\_words}, \texttt{required\_steps}, and style constraints. The answer aliases are included in the scoring criteria and are visible to the GPT-5 judge, but they are not explicitly matched by the deterministic scorer. This hybrid score is used for final evaluation. Inside \textsc{CMI}, intervention utilities are computed using the deterministic task scorer, while the GPT-5 judge is applied only to the final answer for reporting.

We also report task success rate, defined as the fraction of examples whose hybrid task score is at least $0.7$:
\[
    \operatorname{SuccessRate}
    =
    \frac{1}{N}
    \sum_{j=1}^{N}
    \mathbb{I}\left[
        s(y_j,a^\star_j) \geq 0.7
    \right]
\]
Thus, our reported success metric is not exact-match accuracy; it is thresholded hybrid correctness.

\subsection{Memory Selection Metrics}

In addition to answer quality, we evaluate how each method selects memories. Let $\widehat{\mathcal{S}}_j$ be the set of memories selected or treated as exposed for example $j$, $\mathcal{G}_j$ be the set of useful memories, $\mathcal{H}_j$ be the set of harmful or poisoned memories, and $\mathcal{B}_j$ be the set of bad memories. In \textsc{Causal-LoCoMo}, bad memories include both irrelevant and harmful memories. For full-history and summary-memory baselines, these metrics should be interpreted as bookkeeping metrics: the implementation treats the full memory bank as selected for metric computation, even though full-history prompts with past sessions and summary memory prompts with one synthetic summary.

\paragraph{Gold memory recall.}
Gold memory recall measures how many useful memories are selected:
\[
    \operatorname{GoldRecall}
    =
    \frac{1}{N}
    \sum_{j=1}^{N}
    \frac{
        |\widehat{\mathcal{S}}_j \cap \mathcal{G}_j|
    }{
        |\mathcal{G}_j|
    }
\]

\paragraph{Useful memory precision.}
Useful memory precision measures the fraction of selected memories that are useful:
\[
    \operatorname{UsefulPrecision}
    =
    \frac{1}{N}
    \sum_{j=1}^{N}
    \frac{
        |\widehat{\mathcal{S}}_j \cap \mathcal{G}_j|
    }{
        \max(1,|\widehat{\mathcal{S}}_j|)
    }
\]

\paragraph{Useful F1}
Useful F1 is the harmonic mean of useful-memory precision and gold-memory recall:
\[
    \operatorname{UsefulF1}
    =
    \frac{
        2 \cdot \operatorname{UsefulPrecision} \cdot \operatorname{GoldRecall}
    }{
        \max(\epsilon,\operatorname{UsefulPrecision}+\operatorname{GoldRecall})
    }
\]
where $\epsilon$ is a small constant used to avoid division by zero.

\paragraph{Poisoned memory adoption rate.}
The poisoned memory adoption rate measures how often a method selects harmful or poisoned memories when such memories exist:
\[
    \operatorname{PoisonAdoption}
    =
    \frac{1}{N}
    \sum_{j=1}^{N}
    \frac{
        |\widehat{\mathcal{S}}_j \cap \mathcal{H}_j|
    }{
        \max(1,|\mathcal{H}_j|)
    }
\]
This differs from a selected-memory precision metric: the denominator is the number of harmful memories in the example, not the total number of selected memories.

\paragraph{Bad memory rejection rate.}
We also report the rejection rate over all bad memories:
\[
    \operatorname{BadRejection}
    =
    \frac{1}{N}
    \sum_{j=1}^{N}
    \left(
    1 -
    \frac{
        |\widehat{\mathcal{S}}_j \cap \mathcal{B}_j|
    }{
        \max(1,|\mathcal{B}_j|)
    }
    \right)
\]
Since $\mathcal{B}_j$ includes both irrelevant and harmful memories, this metric captures the ability to reject non-useful memories broadly, not only adversarial ones.

\paragraph{Average selected memories.}
Finally, we report the average number of memories selected or exposed by each method:
\[
    \operatorname{AvgMem}
    =
    \frac{1}{N}
    \sum_{j=1}^{N}
    |\widehat{\mathcal{S}}_j|
\]
This helps distinguish methods that perform well by selecting a small number of causally relevant memories from methods that expose the model to large amounts of context.

\subsection{Implementation Details}

All experiments are conducted on the filtered \textsc{Causal-LoCoMo} benchmark described in Section~\ref{sec:dataset}. The benchmark contains 87 examples derived from LoCoMo, spanning temporal memory QA, multi-evidence memory QA, inferential memory QA, and factual memory QA. For each method, we record the generated answer, the selected or exposed memories, the hybrid task score, the thresholded task success indicator, and the memory-selection metrics described above.

The same final response model is used across all methods. In the experimentation, the final response model is GPT-4.1, the judge model is GPT-5, and vector-based retrieval uses \texttt{text-embedding-3-large}. Decoding is run with temperature zero. The answer-generation prompt is held fixed for memory-based methods, and only the supplied context changes. \textsc{CMI} additionally performs pre-answer intervention calls to score no-memory, with-memory, and perturbed-memory conditions before producing its final answer. Reflection memory additionally constructs label-aware reflection strings before retrieval.

\section{Results}
\label{sec:results}

We evaluate all methods on \textsc{Causal-LoCoMo} using both answer-quality and memory-selection metrics. Since the goal of this work is to study whether agents select memories that causally support the current task, task performance alone is insufficient. We therefore report the hybrid task score and success rate together with useful-memory F1, bad-memory rejection, poisoned-memory adoption, and the average number of selected or exposed memories.

\subsection{Main Results}
\label{subsec:main_results}

Table~\ref{tab:main_results} presents the main comparison across all agents. \textsc{CMI} achieves the highest task score ($0.846$) and the highest success rate ($0.816$). It also obtains the strongest useful-memory F1 ($0.875$), near-perfect bad-memory rejection ($0.990$), and zero poisoned-memory adoption. Notably, \textsc{CMI} selects fewer than one memory on average, indicating that its performance does not come from exposing the response model to a larger context.

The strongest non-CMI baselines, reflection memory and vector memory, achieve similar task scores ($0.845$ and $0.839$, respectively), but their memory-selection behavior differs substantially. Both select three memories on average and adopt poisoned memories at much higher rates. Summary memory and full-history prompting expose broader context, but obtain lower task scores, suggesting that larger context alone does not guarantee reliable long-horizon memory use.

\begin{table*}[t]
\centering
\small
\begin{tabular}{lcccccc}
\toprule
\textbf{Agent} & \textbf{Task Score} & \textbf{Success Rate} & \textbf{Useful F1} & \textbf{Bad Mem. Rejection} & \textbf{Poison Adoption} & \textbf{Avg. Memories} \\
\midrule
\textsc{CMI} & \textbf{0.846} & \textbf{0.816} & \textbf{0.875} & \textbf{0.990} & \textbf{0.000} & 0.943 \\
Reflection Memory & 0.845 & 0.793 & 0.486 & 0.557 & 0.540 & 3.000 \\
Vector Memory & 0.839 & 0.782 & 0.501 & 0.566 & 0.609 & 3.000 \\
Graph Memory & 0.824 & 0.759 & 0.469 & 0.550 & 0.586 & 3.000 \\
Summary Memory & 0.723 & 0.586 & 0.308 & 0.000 & 0.621 & 5.644 \\
Full History & 0.515 & 0.218 & 0.308 & 0.000 & 0.621 & 5.644 \\
No Memory & 0.429 & 0.034 & 0.000 & 1.000 & 0.000 & 0.000 \\
\bottomrule
\end{tabular}
\caption{Main performance comparison on \textsc{Causal-LoCoMo}. \textsc{CMI} achieves the best task score, success rate, useful-memory F1, bad-memory rejection, and poisoned-memory adoption.}
\label{tab:main_results}
\end{table*}

\subsection{Accuracy--Robustness Tradeoff}
\label{subsec:accuracy_robustness}

\begin{figure}[htbp]
\centering
\includegraphics[width=0.92\linewidth]{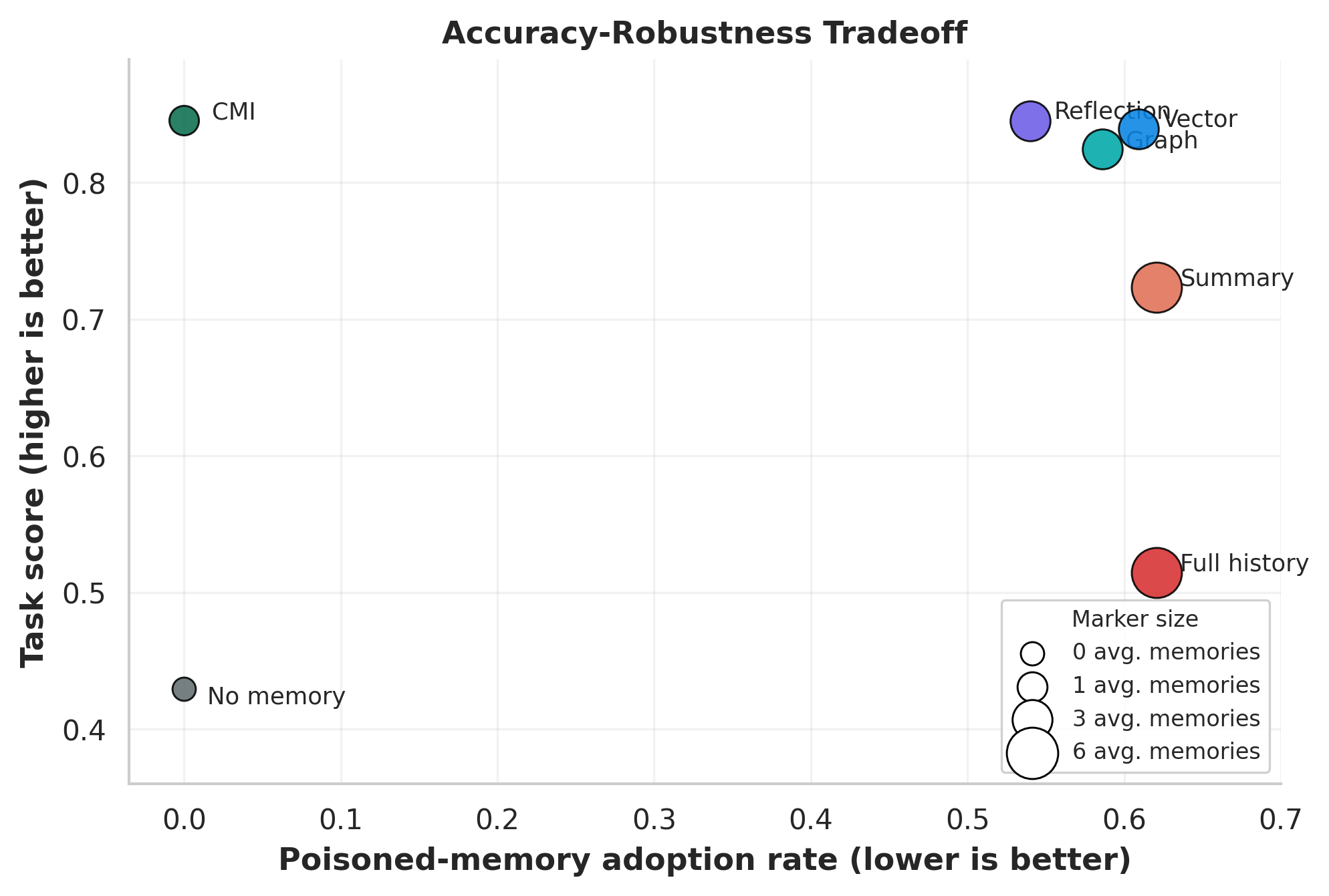}
\caption{Accuracy--robustness tradeoff. Each point corresponds to a memory method; the $x$-axis shows poisoned-memory adoption rate and the $y$-axis shows task score. Marker size indicates the average number of selected or exposed memories.}
\label{fig:accuracy_robustness_tradeoff}
\end{figure}

Figure~\ref{fig:accuracy_robustness_tradeoff} plots each method by task score and poisoned-memory adoption rate. The desired region is the upper-left corner, corresponding to high task performance and low adoption of poisoned memories. \textsc{CMI} occupies this region, with the highest task score and zero poisoned-memory adoption. Vector, graph, and reflection memory achieve competitive task scores, but appear in a higher-risk region because they adopt poisoned memories substantially more often.

\subsection{Performance by Task Family}
\label{subsec:task_family_breakdown}

Table~\ref{tab:task_family} reports task scores by task family. \textsc{CMI} obtains the best score on multi-evidence memory QA ($0.754$) and matches the best score on factual QA ($0.985$). On inferential memory QA, reflection memory obtains the highest score ($0.823$), with \textsc{CMI} close behind ($0.808$). On temporal memory QA, vector memory performs best ($0.925$), followed by reflection memory ($0.918$), graph memory ($0.914$), and \textsc{CMI} ($0.901$). Factual QA contains only two examples, so results for that family should be interpreted cautiously.

\begin{table*}[t]
\centering
\small
\begin{tabular}{lcccc}
\toprule
\textbf{Agent} & \textbf{Temporal QA} & \textbf{Multi-Evidence QA} & \textbf{Inferential QA} & \textbf{Factual QA} \\
\midrule
\textsc{CMI} & 0.901 & \textbf{0.754} & 0.808 & \textbf{0.985} \\
Reflection Memory & 0.918 & 0.738 & \textbf{0.823} & 0.660 \\
Vector Memory & \textbf{0.925} & 0.730 & 0.731 & \textbf{0.985} \\
Graph Memory & 0.914 & 0.725 & 0.731 & 0.660 \\
Summary Memory & 0.822 & 0.600 & 0.583 & \textbf{0.985} \\
Full History & 0.548 & 0.491 & 0.432 & 0.610 \\
No Memory & 0.421 & 0.442 & 0.445 & 0.350 \\
\bottomrule
\end{tabular}
\caption{Task-family breakdown by task score. \textsc{CMI} performs best on multi-evidence QA and remains competitive on temporal and inferential QA. Factual QA contains only two examples.}
\label{tab:task_family}
\end{table*}

\subsection{Causal Utility Diagnostics}
\label{subsec:causal_utility_diagnostics}

\begin{figure}[htbp]
\centering
\includegraphics[width=0.86\linewidth]{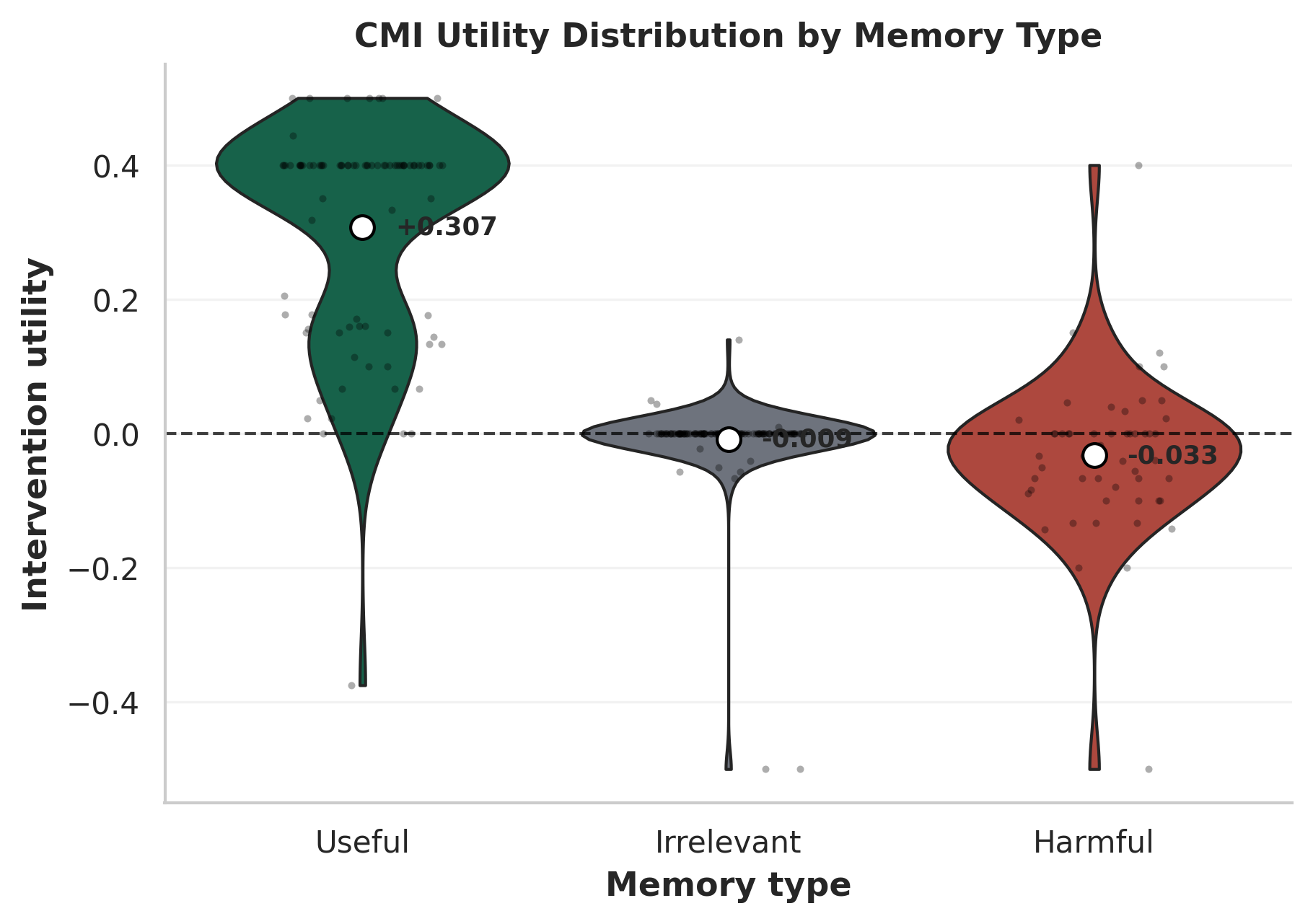}
\caption{Distribution of \textsc{CMI} intervention utility by memory type.}
\label{fig:cmi_utility_distribution}
\end{figure}

We also report the intervention diagnostics produced by \textsc{CMI}. Table~\ref{tab:utility_diagnostics} shows the average intervention utility and stability by memory type, and Figure~\ref{fig:cmi_utility_distribution} shows the corresponding utility distributions. Useful memories have positive average utility ($+0.307$), irrelevant memories have near-zero utility ($-0.009$), and harmful memories have negative utility ($-0.033$). Average stability is small across memory types, with useful memories having the highest value ($+0.027$).

\begin{table}[htb]
\centering
\small
\begin{tabular}{lcc}
\toprule
\textbf{Memory Type} & \textbf{Average Utility} & \textbf{Average Stability} \\
\midrule
Useful memories & $+0.307$ & $+0.027$ \\
Irrelevant memories & $-0.009$ & $+0.000$ \\
Harmful memories & $-0.033$ & $+0.001$ \\
\bottomrule
\end{tabular}
\caption{Causal utility diagnostics for \textsc{CMI}. Utility is computed as the change in deterministic task score between the with-memory and no-memory conditions.}
\label{tab:utility_diagnostics}
\end{table}

\section{Analysis and Discussion}
\label{sec:analysis_discussion}

The results suggest that the main advantage of \textsc{CMI} comes from changing the criterion used for memory selection. Standard memory systems select context using semantic similarity, graph proximity, reflection-style representations, summaries, or full conversation histories. These mechanisms can surface memories that appear relevant to the current request, but they do not directly test whether the selected memory improves the final answer. \textsc{CMI} instead evaluates candidate memories through intervention: a memory is selected only if it improves the deterministic task score relative to the no-memory condition and remains stable under perturbation. This explains why \textsc{CMI} achieves the best overall task score while selecting fewer than one memory on average. Its improvement does not come from increasing context size; rather, it comes from filtering the memory bank toward memories with positive estimated causal effect. The accuracy--robustness tradeoff further supports this interpretation: \textsc{CMI} lies in the desirable high-score, low-poison-adoption region, whereas vector, graph, and reflection memory obtain competitive task scores but adopt poisoned memories at substantially higher rates.

The behavior of the baselines reveals why semantic or broad-context memory access is insufficient for reliable long-horizon agents. Vector memory performs strongly on temporal questions, likely because many temporal tasks contain explicit lexical or semantic cues that make the relevant memory easy to retrieve. Reflection and graph memory also remain competitive because their representations preserve useful task-related signals. However, these same mechanisms are vulnerable when harmful memories are designed to be plausible and topically close to the query. A poisoned memory may mention the correct person, event, or time period while introducing an incorrect answer, making it attractive to similarity-based or keyword-based selectors. This explains why these baselines can achieve high task scores while still showing weak useful-memory F1 and high poisoned-memory adoption. In contrast, full-history prompting and summary memory underperform despite exposing more context. Full-history prompting forces the model to search through a large amount of conversational information, which can dilute the relevant evidence and introduce competing details. Summary memory reduces context length, but may remove fine-grained information such as exact dates, entity distinctions, or multi-evidence dependencies. These results indicate that the central bottleneck is not memory availability, but selective and reliable memory use.

The causal utility diagnostics provide direct evidence for the mechanism behind \textsc{CMI}. Useful memories have strongly positive average utility, irrelevant memories are centered near zero, and harmful memories have negative average utility. This separation shows that intervention scores capture meaningful differences between memory types: useful memories tend to improve the model's answer, irrelevant memories usually have little effect, and harmful memories tend to degrade performance. The distributional result also explains why \textsc{CMI} performs especially well on multi-evidence questions, where the model must identify supporting memories while ignoring distractors. At the same time, the task-family breakdown shows that \textsc{CMI} is not uniformly best on every category; vector memory slightly outperforms it on temporal QA, where surface cues may be sufficient for retrieval. Thus, the key contribution of \textsc{CMI} is not that intervention-based selection always dominates semantic retrieval on every task, but that it provides a stronger overall accuracy--robustness tradeoff. More broadly, these findings suggest that long-term memory for LLM agents should be evaluated not only by final answer quality, but also by whether the selected memories are causally useful, stable, and safe to expose to the model.

\section{Limitations and Ethical Considerations}
\label{sec:limitations}

This work studies \textsc{CMI} in a controlled, annotated memory-selection setting, and the results should be interpreted accordingly. In the current implementation, memory entries may include role annotations such as useful, irrelevant, or harmful, and some methods use these annotations during selection. Thus, our experiments evaluate the value of explicit causal-memory structure rather than proving that agents can infer causal roles from raw memories alone. A fully deployable version of \textsc{CMI} would require replacing gold memory-role labels with predicted causal-role estimates and evaluating those predictions under realistic ambiguity, distribution shift, and adversarial conditions. In addition, \textsc{Causal-LoCoMo} is relatively small: the reported experiments use 87 filtered examples derived from LoCoMo~\cite{maharana2024locomo}, concentrated mainly in temporal, multi-evidence, and inferential memory QA. Although we apply LLM-assisted construction, deterministic filtering, schema validation, and leakage checks, the benchmark does not cover the full range of memory failures that may occur in deployed agents, including stale user preferences, incorrect summaries, entity-linking errors, conflicting updates, and long-term memory drift. The harmful memories are also synthetic adversarial insertions; they are useful for controlled robustness evaluation, but may not fully capture naturally occurring or human-crafted memory corruptions.

Our evaluation also has methodological and computational limitations. The final task score combines a deterministic scorer with a GPT-5 judge, which is more flexible than exact matching but introduces dependence on the judge model's calibration, prompt sensitivity, and possible biases. Future work should include larger benchmarks, label-free memory settings, human-authored memory corruptions, independent human evaluation, and judge-model ablations. \textsc{CMI} also incurs additional inference cost because it performs pre-answer intervention calls over no-memory, with-memory, and perturbed-memory conditions before producing the final response. This cost is justified in reliability-sensitive settings, but it increases latency and token usage relative to simple retrieval. Practical deployments may need to cache memory utilities, restrict intervention scoring to high-risk queries, or use smaller verifier models for the intervention step.

Long-term memory systems also raise important ethical concerns because they store and reuse information about users across interactions. Memories may contain sensitive personal attributes, preferences, health information, relationships, or identity-related disclosures. Better memory selection can improve personalization and reliability, but it can also increase privacy risk if deployed without consent, access control, retention limits, provenance tracking, and user-facing mechanisms to inspect, correct, or delete memories. Causal selection itself involves tradeoffs: overly aggressive filtering may suppress sensitive but relevant context, while overly permissive filtering may reinforce stale, incorrect, or biased memories. Finally, synthetic harmful memories are useful for robustness evaluation, but similar constructions could be misused to attack memory-augmented agents~\cite{liu2023promptinjection,zou2025poisonedrag,srivastava2025memorygraft}. We therefore view \textsc{CMI} as one component of a broader responsible memory architecture, which should include data minimization, user consent, memory expiration, uncertainty estimation, provenance tracking, and periodic audits for privacy leakage, bias, and harmful memory persistence.

\section{Conclusion}

We introduced \textsc{Causal Memory Intervention} (\textsc{CMI}), a causal intervention-based memory-selection technique for long-horizon LLM agents. Rather than retrieving memories solely because they are semantically similar to the current request, \textsc{CMI} estimates whether candidate memories improve the model's answer under controlled intervention conditions and selects only memories with positive and stable causal effect. To evaluate this setting, we constructed \textsc{Causal-LoCoMo}, a causally annotated benchmark derived from long conversational data, and compared \textsc{CMI} against vector, graph, reflection, summary, full-history, and no-memory baselines. Our experiments show that \textsc{CMI} achieves the strongest overall balance between answer quality and memory-selection reliability, obtaining the best task score and useful-memory F1 while substantially reducing bad-memory selection and eliminating poisoned-memory adoption in the reported setting. These findings suggest that reliable long-term memory for LLM agents requires more than access to larger context windows or semantically relevant memories. It requires selecting memories according to their causal effect on the current answer. We hope this work encourages future memory-augmented agent systems to treat memory selection as a causal decision problem, with greater emphasis on robustness, provenance, and safe exclusion of misleading context.

\bibliography{example_paper}
\bibliographystyle{icml2026}

\end{document}